\documentclass[conference]{IEEEtran}
\usepackage{amsmath}
\usepackage{hyperref}
\usepackage{amssymb}
\usepackage{graphicx}
\usepackage{ragged2e}
\usepackage{amsmath}
\usepackage{array}
\usepackage{gensymb}
\usepackage{xcolor}
\usepackage{soul}
\usepackage[linesnumbered,ruled,vlined]{algorithm2e}

\usepackage{comment}

\newcolumntype{M}[1]{>{\centering\arraybackslash}m{#1}}

\usepackage{caption}

\begin{document}
\title{Edge-Enabled Collaborative Object Detection for Real-Time Multi-Vehicle Perception}

%\author{\\ \\ \\ \\ \\}

\author{\IEEEauthorblockN{Everett Richards}
\IEEEauthorblockA{San Diego State University\\
San Diego, California, USA\\
Email: ehrichards@sdsu.edu}
\and
\IEEEauthorblockN{Bipul Thapa}
\IEEEauthorblockA{University of Delaware\\
Newark, Delaware, USA\\
Email: bipul@udel.edu}
\and
\IEEEauthorblockN{Lena Mashayekhy}
\IEEEauthorblockA{University of Delaware\\
Newark, Delaware, USA\\
Email: mlena@udel.edu}}

\maketitle

\begin{abstract}
Accurate and reliable object detection is critical for ensuring the safety and efficiency of Connected Autonomous Vehicles (CAVs). 
Traditional on-board perception systems have limited accuracy due to occlusions and blind spots, while cloud-based solutions introduce significant latency, making them unsuitable for real-time processing demands required for autonomous driving in dynamic environments.
To address these challenges, we introduce an innovative framework, Edge-Enabled Collaborative Object Detection (ECOD) for CAVs, that leverages edge computing and multi-CAV collaboration for real-time, multi-perspective object detection. Our ECOD framework integrates two key algorithms: Perceptive Aggregation and Collaborative Estimation (PACE) and Variable Object Tally and Evaluation (VOTE). PACE aggregates detection data from multiple CAVs on an edge server to enhance perception in scenarios where individual CAVs have limited visibility. VOTE utilizes a consensus-based voting mechanism to improve the accuracy of object classification by integrating data from multiple CAVs. Both algorithms are designed at the edge to operate in real-time, ensuring low-latency and reliable decision-making for CAVs. We develop a hardware-based controlled testbed consisting of camera-equipped robotic CAVs and an edge server to evaluate the efficacy of our framework. Our experimental results demonstrate the significant benefits of ECOD in terms of improved object classification accuracy, outperforming traditional single-perspective onboard approaches by up to~75\%, while ensuring low-latency, edge-driven real-time processing. 
This research highlights the potential of edge computing to enhance collaborative perception for latency-sensitive autonomous systems.
\end{abstract}

\IEEEpeerreviewmaketitle
\section{Introduction}
Autonomous Vehicles (AVs) have seen steady growth in recent years, with clear indications of rapid expansion in the coming decades. According to Litman~\cite{litman2023},  operational AVs are expected to be commercially available by~2030, and could become affordable and widespread between~2040 and~2060. Additionally, Moody~\cite{moody2020} reports that nearly half of those surveyed perceives AVs as ``very" or ``somewhat" safe. Despite these positive projections, significant safety concerns persist, hindering the broader adoption of AV technology.

A major challenge in ensuring AV safety lies in object detection and situational awareness. Current AV systems usually rely on onboard sensors (e.g., cameras, LiDAR, radar) to detect and classify objects, but they are inherently limited by occlusions, blind spots, and sensor noise, which can cause them to misinterpret their surroundings~\cite{chu2025occlusion, xiao2023overcoming}. 
These limitations become particularly pronounced in complex and dynamic environments, such as parking lots and intersections, where poor visibility or unexpected object movements can lead to misclassifications and collisions. Although advancements in computer vision and LiDAR technologies have improved detection capabilities, these systems remain insufficient for ensuring the high level of accuracy and reliability required for consistent safety. The reliance on isolated sensor data from a single AV limits the system's ability to generate a comprehensive understanding of its surroundings, particularly when visibility is obstructed. 

The real-world implications of these shortcomings are demonstrated by the National Highway Traffic Safety Administration (NHTSA). The NHTSA's ongoing investigation highlights nearly 1,000 accidents involving Tesla's autopilot features between~2018 and~2023, with over two dozen fatalities~\cite{nhtsa}. Many of these accidents were due to failures in object classification, such as a notable case where a Tesla vehicle misidentified a truck as a cloud in the~sky. The report further reveals that approximately~20\% of these accidents occurred with stationary objects, highlighting the limitations of single-vehicle perception and the risks present even in low-speed environments. These failures emphasize the critical need for enhanced object detection and decision-making capabilities in AVs to ensure their safe operation.

A major factor contributing to these limitations is the sensing modality used in many commercial AVs. While advanced perception systems often integrate LiDAR and radar, camera-only sensing remains widespread due to its lower cost and energy efficiency. For instance, Tesla has eliminated LiDAR and radar from its Autopilot and Full Self-Driving platforms, relying exclusively on camera-based inputs~\cite{teslaVision2022}. This shift is driven by the declining cost of high-resolution optical sensors and the scalability of vision-based deep learning. However, relying solely on a single vehicle's camera can result in blind spots and misclassifications, particularly in complex and occluded environments. %To address this shortcomings, our ECOD framework emphasizes optical perception, while remaining modular enough to incorporate additional modalities in future extensions.

To mitigate the limitations of single-vehicle camera-only sensing, researchers have emphasized the importance of collaborative perception among multiple Connected Autonomous Vehicles (CAVs), where vehicles share sensory data to improve detection accuracy~\cite{shetty2021}. This multi-CAV cooperation can significantly enhance situational awareness by fusing data from different vantage points, therefore reducing occlusion-related errors, and improving overall safety.
%%mitigate sensor occlusion and improve overall safety and reduce the risk of fatal accidents.
%Considering these safety concerns, Shetty et al. \cite{shetty2021} highlight the importance of continuous collaboration among CAVs to enhance safety and reduce the risk of fatal accidents. 
%%This paradigm, , is particularly critical in situations with poor visibility or where CAVs may make unexpected decisions. %Many of the collisions outlined in the NHTSA report could have been avoided by forming collaborative decisions using multiple vehicles' perspectives.
However, existing collaborative frameworks predominantly rely on cloud computing, which introduces high latency and bandwidth constraints. While cloud-based systems are suitable for applications with moderate latency tolerance, they are inadequate for the real-time requirements of autonomous driving, where even milliseconds of delay can compromise safety~\cite{lu2023vehicle}. 

\begin{figure}[tbp]
    \centering
    \includegraphics[width=0.95\linewidth]{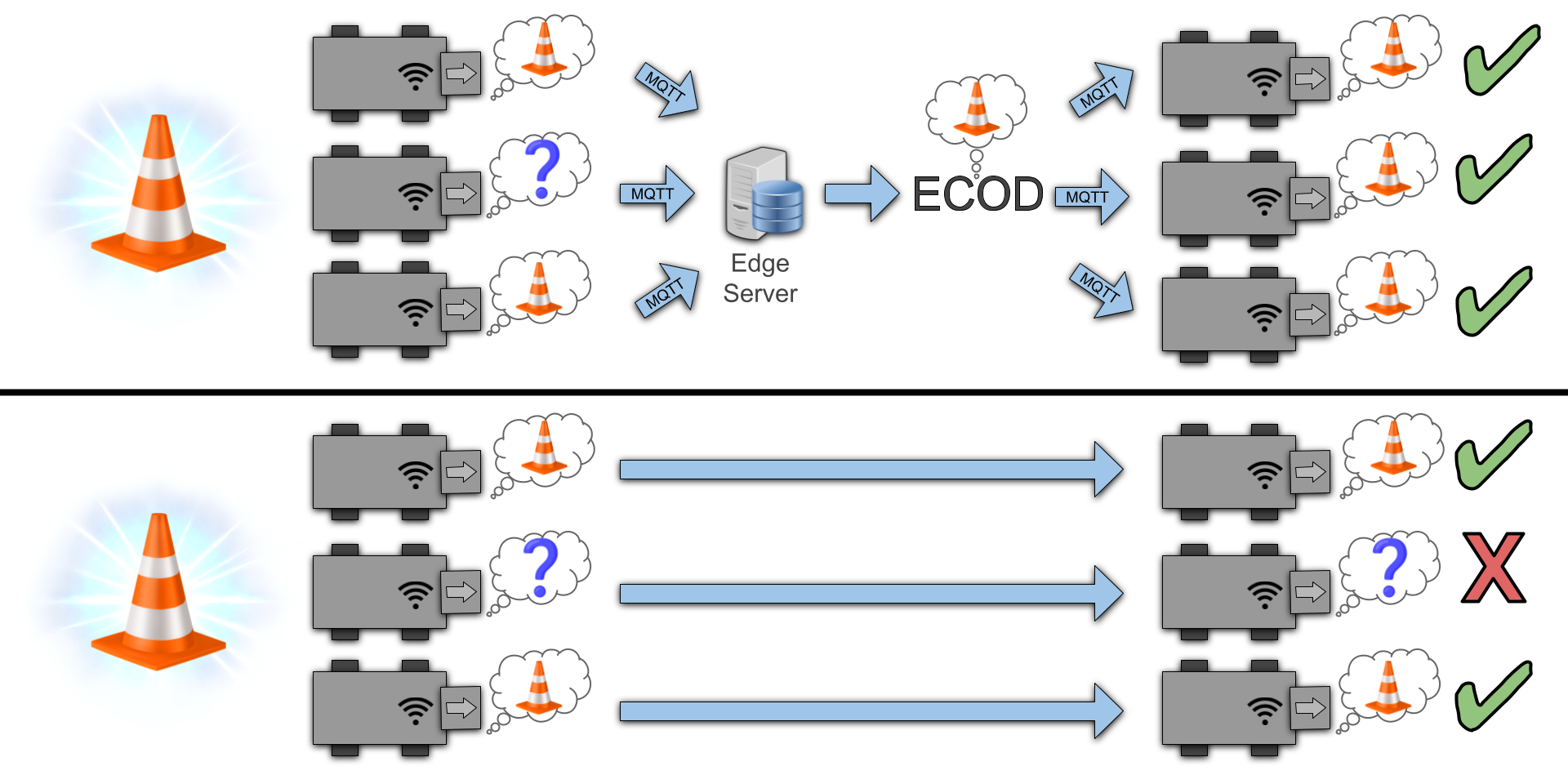}
    \caption{Comparison of Collaborative (top) and Individual (bottom) Object Detection}
    \label{fig:collab_vs_indiv}
    \vspace*{-0.2cm}
\end{figure}

%Building on this need for real-time collaboration, traditional multi-agent systems have often relied on cloud computing. While cloud computing is suitable for applications with moderate latency tolerance, such as remote file storage and various web services, it falls short of the low-latency demands crucial for CAV systems. 

To support the low-latency requirements of CAVs, a transition to edge computing is essential. Edge computing offers a distributed computing framework that brings computation closer to the data source at the network edge, reducing latency and bandwidth costs, and enhancing privacy~\cite{thapa2024latency, maleki2024qos}. 
Recent studies have explored edge-assisted cooperative perception, but many rely on raw or feature-level data fusion, which is computationally intensive and bandwidth-heavy. Furthermore, existing methods lack real-world validation, as they are often evaluated in simulated environments rather than on physical testbeds.

Building on the advantages of edge computing, we propose  Edge-Enabled Collaborative Object Detection (ECOD), a novel framework that leverages edge computing and multi-CAV collaboration for real-time, multi-perspective object detection (illustrated in Figure~\ref{fig:collab_vs_indiv}).  
%This framework facilitates communication between CAVs through edge servers, enabling collaborative object classification by aggregating data from multiple sources. The edge server collects detected object labeling data from each AV, applies a voting approach to determine the most accurate classification, and then publishes the consensus result back to the vehicles.
%that leverages edge computing to enable accurate real-time collaborative decision-making among CAVs for object classification. 
ECOD includes two distinct algorithms: Perceptive Aggregation and Collaborative Estimation (PACE) and Variable Object Tally and Evaluation (VOTE). PACE aggregates object detection data from multiple CAVs at the edge server to enhance perception by providing a comprehensive view of the environment, especially in situations where individual CAVs have limited visibility. %only a few CAVs have a direct and unobstructed view of an object, to create a collaborative estimation of the object. 
VOTE enhances the accuracy of object classification through a consensus-based voting mechanism that integrates data from multiple CAVs while accounting for CAV reputation and object visibility constraints. 
%The second, the VOTE algorithm, involves collecting labeled data from each CAV simultaneously, using a voting mechanism at the edge server to compare each label's viability, selecting the most likely labels, and then distributing the consensus result back to the CAVs.
By integrating these algorithms, the ECOD framework enables more precise situational awareness while addressing the limitations of single-CAV perception. 

We evaluate the effectiveness of the ECOD framework by deploying a hardware-based testbed consisting of four small camera-equipped robotic CAVs and an edge server. 
We consider two usecases: intersection mapping and parking lot tracking. 
We consider various traffic scenarios to assess ECOD's collaborative performance. 
The results demonstrate a significant improvement in decision accuracy compared to single-CAV onboard systems, with ECOD outperforming traditional onboard approaches by up to 75\%. 
ECOD reduces latency by leveraging edge-based processing, ensuring real-time collaborative decision-making. In addition, it enhances scalability, making it suitable for multi-CAV networks in real-world autonomous systems. 
By integrating edge computing with multi-CAV collaboration, ECOD enables low-latency, real-time perception, contributing to the broader deployment of edge-assisted autonomous driving technologies.

The rest of the paper is organized as follows. In Section~\ref{related_work}, we review existing work in this domain. In Section~\ref{ECOD_approach}, we outline our framework and describe both PACE and VOTE. In Section~\ref{experimentation}, we present the experimental setup and evaluate the results. 
Section~\ref{conclusion} summarizes our findings and outlines potential future research directions.

\begin{table*}[t] 
\centering
\renewcommand{\arraystretch}{1.20}
\caption{Comparison With Existing Research} \vspace*{-0.1cm}
\label{table:refs}
\begin{tabular}{ >{\RaggedRight}m{2.5cm}|M{0.5cm}|M{0.5cm}|M{0.5cm}|M{0.5cm}|M{0.5cm}|M{0.5cm}|M{0.5cm}|M{0.7cm} }

&\multicolumn{8}{c}{\bf Studies} \\
\hline %Research Topic
 & \cite{yee2018} & \cite{he2023} & \cite{li2018} & \cite{d'ortona2022} & \cite{luo2023edgecooper} & \cite{liu2021edgesharing} & \cite{song2022edge} & ECOD \\
\hline
Collaborative Perception & & & \checkmark & & \checkmark & \checkmark & \checkmark & \checkmark\\
\hline
Edge Computing & & \checkmark & & & \checkmark & \checkmark & \checkmark & \checkmark\\
\hline
MQTT Protocol & & & & \checkmark & & & & \checkmark\\
\hline
%Voting algorithm & & & & \checkmark & & \checkmark\\
%\hline
%Parking lots & & & \checkmark & & & \checkmark\\
%\hline
%Latency-aware experimentation & & \checkmark & \checkmark & \checkmark\\
%\hline
V2N Interaction & & \checkmark & \checkmark & \checkmark & \checkmark & \checkmark & \checkmark & \checkmark\\ 
\hline

%Optimized for Edge / Low Latency & & \checkmark & & \checkmark & & \checkmark & & \checkmark\\
%\hline

Physical Testbed & \checkmark & \checkmark & \checkmark & \checkmark & & & & \checkmark\\

\end{tabular}

\vspace{-0.2cm}
\end{table*}

\section {Related Work}
\label{related_work}
In response to growing interest in intelligent and autonomous vehicles, Vehicle-to-Vehicle (V2V), Vehicle-to-Infrastructure (V2I), Vehicle-to-Network (V2N), and Vehicle-to-Everything (V2X) communication protocols have been studied extensively in the past few years~\cite{malik2021}. These protocols are crucial for enabling real-time situational awareness and improving decision-making in autonomous systems, which are often hindered by the limitations of single-vehicle, single-perspective object classification techniques. In this section, we review key research papers that have made substantial and relevant contributions in collaborative perception, edge computing for CAVs, and networking approaches that form the foundation for our work.

\vspace*{0.2cm}
\noindent \emph{Networking and V2X-Based Approaches.} Several studies have explored networking solutions for CAVs. Yee et al.~\cite{yee2018} explored collaborative perception using a single vehicle with two cameras to mimic multiple viewpoints to provide a framework for V2X object detection. 
However, this study does not achieve true multi-vehicle perception, as its setup is limited to a single-vehicle testbed with two mounted cameras (multi-sensor fusion) rather than V2V data exchange. 
Li et al.~\cite{li2018} introduced a collaborative paradigm that leverages LiDAR data from autonomous vehicles to generate real-time 3D maps of multi-story parking garages, but focus on static environments. 
D'Ortona et al.~\cite{d'ortona2022} expanded upon existing inter-vehicle communication solutions by proposing the use of the MQTT protocol for communication between vehicles and vulnerable road users (such as pedestrians and cyclists) using Bluetooth Low Energy (BLE). Their approach assumes that all road users are equipped with BLE-enabled devices, limiting its scalability.  
Other related work involving MQTT in V2X systems includes approaches by Shin and Jeon~\cite{shin2024mqtree}, Affia and Matulevičius~\cite{affia2021}, and Hadded et al.~\cite{hadded2022assessment}. However, these studies each explore the applications of MQTT in solving specific problems in autonomous vehicles (distributed software updates, traffic light perception, and the impact of cybersecurity threats on MQTT in autonomous vehicle systems, respectively). 
Moreover, these approaches did not leverage the benefits of edge computing.

\vspace*{0.2cm}
\noindent \emph{Collaborative Perception in CAVs. } The need for robust collaborative perception spans a variety of domains, from battlefield IoT systems~\cite{sadik2024collaborative} to connected autonomous vehicles. Prior research has explored multi-agent perception using various levels of data fusion. 
Luo et al.~\cite{luo2023edgecooper} proposed EdgeCooper,  a LiDAR-based cooperative perception framework, where raw LiDAR data from multiple vehicles is transmitted to an edge server for fusion (raw-level fusion). However, this approach is bandwidth-intensive and requires extensive edge processing, making it less practical.  
Similarly, Liu et al.~\cite{liu2021edgesharing} presented EdgeSharing, which constructs a 3D feature map to facilitate collaborative localization and object sharing in urban environments, while Song et al.~\cite{song2022edge} focuses on sensor noise estimation and fusion in vehicular communication networks. These methods employ feature-level fusion, reduce bandwidth usage compared to raw fusion but still require increases computational overhead and demands significant network resources.

Zhang et al.~\cite{zhang2021emp} developed EMP, utilizing LiDAR-based sensing with an object-level fusion, which reduces data transmission requirements by exchanging only detected objects. While object-level fusion is more efficient, EMP primarily relies on LiDAR-based sensing, which may not generalize well to diverse sensor modalities such as cameras and radar. Additionally, EMP lacks mechanisms to handle inconsistencies in multi-vehicle detections. 
Wang et al.~\cite{wang2020v2vnet} proposed V2VNet, which employs  intermediate feature fusion and 3D convolution to aggregate LiDAR-based data from nearby vehicles. Lin et al.~\cite{lin2024v2vformer} introduced V2VFormer and Xu et al.~\cite{xu2022v2x} developed V2X-ViT, both utilizing transformer-based architectures to capture spatial dependencies and integrate multi-vehicle perspectives. While these approaches achieve high accuracy in simulated urban environments using platforms such as SUMO, CARLA, and NS3, they often require high bandwidth for feature transmission and lack physical testbed validation. In contrast, ECOD performs lightweight object-level fusion on real hardware, enabling low-latency inference with practical feasibility. Our work complements these state-of-the-art methods by providing a modular, testbed-validated edge-based perception framework suitable for deployment in constrained and heterogeneous environments.

%\textcolor{red}{\st{Moreover, these studies rely on simulation-based testing using frameworks like SUMO, CARLA, and NS3, rather than practical hardware-based testbeds, limiting their applicability in dynamic traffic environments.}

\vspace*{0.2cm}
\noindent \emph{Edge Computing for Autonomous Systems. } Edge computing is increasingly leveraged to support low-latency, real-time decision-making~\cite{zhang2025ftformer,bornholdt2025software}. Prior work on edge-based resource management and offloading in vehicular systems has addressed challenges such as privacy~\cite{Ma:ucc19} and computation placement~\cite{Bhatta:CloudCom19}. 
He et al.~\cite{he2023} proposed an edge-enabled C-V2X (cellular vehicle-to-everything) communication framework, where each vehicle frequently broadcasts motion data, such as speed and heading, to nearby vehicles. 
However, they assumed that all vehicles in the vicinity are equipped with V2X technology such as a configurable digital BIOS and intelligent transponders, which limits the framework's applicability in current mixed-vehicle environments. Moreover, their framework does not support collaborative object detection, focusing instead on individual vehicle motion awareness. 
Other edge-based studies, such as EdgeCooper~\cite{luo2023edgecooper} and EdgeSharing~\cite{liu2021edgesharing}, utilize edge servers for processing, but primarily rely on feature-heavy or raw sensor fusion, which can introduce latency bottlenecks. ECOD addresses these challenges by performing object-level fusion at the edge, reducing network overhead while maintaining real-time performance.
Additionally, existing edge-based frameworks do not explicitly address communication delays or object detection inconsistencies across multiple vehicles. ECOD's VOTE algorithm introduces a reputation-based voting mechanism to mitigate discrepancies in classification, ensuring more robust consensus-driven perception.

Despite progress in collaborative perception and edge computing, existing methods suffer from high bandwidth consumption, lack of real-world validation, and computational inefficiencies. Our research addresses addresses these limitations. 
Table~\ref{table:refs} summarizes the key differences between ECOD and existing methods, 
demonstrating its practical feasibility for edge-assisted collaborative perception.

\section{Edge-Enabled Collaborative Object Detection (ECOD)}
\label{ECOD_approach}
In this section, we introduce the ECOD framework, designed to enable groups of CAVs to  collaborate in real-time for multi-perspective object detection using edge computing. ECOD enhances perception accuracy, mitigates sensor occlusions and blind spots, and reduces latency by fusing object detection results from multiple CAVs. Unlike single-CAV perception, which is limited by localized sensor coverage, ECOD integrates collaborative intelligence to improve detection robustness in dynamic environments.

At its core, ECOD provides a pipeline for reliable low-latency data transmission between CAVs and an edge server, employing  two key algorithms for determining verdicts: Perceptive Aggregation and Collaborative Estimation (PACE) and Variable Object Tally and Evaluation (VOTE). These algorithms enable robust and scalable object detection, ensuring accurate classification decisions even in challenging urban environments.

\subsection{System Overview}
ECOD consists of a two-layer architecture comprising CAVs and an Edge Server. The system operates as follows:
\begin{itemize}
\item CAV Layer: Each CAV is equipped with onboard sensors (e.g., cameras, LiDAR) that detect objects in its surroundings and generate preliminary object classifications.

\item Edge Layer: The detected object data is transmitted to a nearby edge server, which aggregates multi-CAV detections, matches objects, and applies collaborative filtering techniques. The edge server processes fused data using the PACE and VOTE algorithms to reach consensus on object classification and positioning, reducing errors caused by individual sensor limitations. 
\end{itemize}
Unlike cloud-based approaches, ECOD ensures low-latency decision-making by processing data at the network edge, enabling real-time object detection for dynamic traffic settings.

The networking module forms the backbone of the ECOD framework, providing the infrastructure for continuous data exchange between CAVs and the edge server. 
To ensure efficient communication, CAVs connect to the edge server via a dedicated Wi-Fi network hosted by the server itself. This configuration ensures secure and stable high-bandwidth data transmission. 
To facilitate real-time communication, ECOD utilizes the MQTT protocol, which follows a publisher/subscriber model. This allows multiple CAVs to simultaneously transmit data to the edge server. In addition, the server uses MQTT when returning its classification verdicts to all CAVs in the area. Data is exchanged at the object level, as opposed to feature-level or raw-level, in order to minimize latency and reduce the computational load on the edge server.

\subsection{Perceptive Aggregation and Collaborative Estimation (PACE)}
The PACE algorithm is designed for multi-CAV cooperative perception, particularly in scenarios where individual CAVs may not have a direct line of sight to the same object, such as in parking lots or urban intersections. 
By leveraging edge computing, PACE allows CAVs to share detected object information with an edge server, which then compiles these detections into a unified global perception map. This ensures that CAVs can perceive objects beyond their direct line of sight, enhancing situational awareness and navigation efficiency. 
PACE follows a two-stage process, involving both CAV-level detection and edge-level object matching and mapping.

\vspace*{0.2cm}
\noindent{\textbf{{PACE: CAV Component}.}}
Each CAV operates a local PACE client algorithm (shown in Algorithm~\ref{alg:PACE_Client}) to independently detects objects and estimates their properties. 
The client algorithm runs locally on a V2N-enabled CAV which continuously scans its surroundings using onboard sensors and a computer vision model~$\mathcal{M}$.
To enhance perception beyond its immediate field of view, the CAV subscribes to the global object map (global\_detections) via MQTT to receive the aggregated global perception map from the edge server (line~1). 
The CAV processes incoming camera data to detect objects using~$\mathcal{M}$, which assigns each detected object~$\omega$ a classification label~$l_\omega$, a confidence score~$c_\omega$, and a bounding box with Cartesian coordinates~$b_\omega$=$(x_{\omega}^{\min}, y_{\omega}^{\min}, x_{\omega}^{\max}, y_{\omega}^{\max})$ (line~5). Using this bounding box, the CAV estimates the object's relative position in its own frame of reference by considering the camera’s field of view, angle per pixel~$\gamma$, and the estimated physical size~$s_\omega$ of the object.

\textfloatsep=0.4cm

\begin{algorithm}[t]
\KwData{Camera input (Angle per Pixel $\gamma$, field of view), local computer vision model $\mathcal{M}$, GPS ($x_{v}, y_{v}, \theta_{v}$) for CAV $v \in V$}
\KwResult{Published object detections with positions and labels}

\texttt{mqtt.subscribe}(``global\_detections''); /* CAV subscribes to receive final object map from the edge */\;
%\tcp{CAV subscribes to receive global object map}
        
$\Omega_v \gets \emptyset $; /* Object detection list */ 

\While{true}{
    /* Detect objects using the local model */
    
    $\mathcal{O} \gets \mathcal{M}.${\texttt{detect()}}; /* Get object labels, confidence scores, and bounding box params. */

    /* Compute global position for each detected object */
    
    \ForEach{object $\omega \in \mathcal{O}$}{

        $w_\omega \gets x_{\omega}^{\max} - x_{\omega}^{\min} $; /* Bounding box width */
        
        $s_\omega \gets \omega.size$; /* Estimated true object size */
    
        $\alpha \gets \gamma \times w_\omega$\;

        $d_\omega \gets s_\omega / \tan(\alpha)$; /* distance */ 
        
        $x_{\text{center}} \gets x_{\omega}^{\min} + \frac{1}{2}w_\omega$;\

        /* Compute relative angle and global coordinates */
        
        $\theta_\omega \gets \theta_v - (\gamma \times x_{\text{center}})$\; 
        $x_{\omega} \gets x_{v} + d_\omega \times \cos(\theta_\omega)$\;
        $y_{\omega} \gets y_{v} + d_\omega \times \sin(\theta_\omega)$\;
          
        /* Store object position */
        
        $\omega.position \gets (x_{\omega}, y_{\omega})$\;
        $\Omega_v$.\texttt{append($\omega$)}\;
    }    

    /* Publish detections at regular intervals */
    
    \If{$t- t_p \geq \tau $}{
        \texttt{mqtt.publish}(``detections'', $\Omega_v)$\;
        
        /* Clear object list after publishing */
        
        $\Omega_v$.\texttt{clear()};\

        $t_p \gets t$; /* Update timestamp */\
        
    }
     
}
\caption{PACE CAV Pseudocode}
\label{alg:PACE_Client}
\end{algorithm}

Then, the estimated object position is transformed into global coordinates (i.e., the absolute position of the object) using the CAV's own global coordinates and its orientation angle~$\theta_v$ (obtained via GPS or an experimental localization system), which represents the heading of the CAV in the global frame (lines~8-16). 
The CAV maintains a dynamic list~$\Omega_v$ of detected objects, including their global positions, classification labels, and confidence values (lines~18-19), and it publishes this data to the edge server every~$\tau$ seconds  via MQTT (detections) (lines~21-25). This continuous perception process is repeated in real-time, ensuring that the edge server receives frequent updates from all CAVs, allowing for multi-perspective aggregation. 

\vspace*{0.2cm}
\noindent{\textbf{PACE: Edge Server Component.}}
The edge server component of PACE (shown in Algorithm~\ref{alg:PACE_Server}) acts as a fusion center, consolidating object detections and generating a refined global perception map. 
It manages interactions with multiple CAVs simultaneously by processing multi-CAV incoming data through the following steps. 
The server continuously listens for object detections from CAVs by subscribing to the detections topic via MQTT (line~1). 
As CAVs publish their detected objects, the server receives and updates a list of reported detections from all actively connected CAVs (lines~3-5). Every~$\tau$ seconds, the server processes the latest object detections and consolidates them into a unified detection list~$\Omega_{all}$, which contains object labels, confidence scores, and estimated positions (lines~8-11). 
To construct the global map, the edge server associates these detected objects with real locations~$\rho \in \mathcal{R}$ by comparing the reported positions (lines~13-30). Objects reported within a distance threshold~$\delta$ of a given location~$\rho$ are grouped together into~$\mathcal{O}_\rho$ (lines~16-18). This is to reduce sensor-based inaccuracies (positional noise) and aligning observations from multiple viewpoints. For the detected object located near~$\rho$, the edge server assigns the highest confidence label, the confidence score (greater weight to detections with higher confidence values), and the estimated position based on all contributing CAVs (lines~19-30). 
The edge server compiles this information into the global object map,~$object\_map$, and publishes it via MQTT to the global\_detections topic, ensuring that all CAVs receive the updated global perception data in real-time (lines~31-34). 

%calculates an aggregated confidence score (giving more weight to higher-confidence detections), 

\begin{algorithm}[t]
\KwData{List of actively-connected CAVs ($V$), global object location map ($\mathcal{R}$), distance threshold ($\delta$), update interval ($\tau$)}
\KwResult{Global object map with assigned labels}

\texttt{mqtt.subscribe}(``detections''); /* Edge Server subscribes to CAV detections */

\While{true}{
    /* Receive and update list of client-mapped data from CAVs */
    
    \ForEach{CAV $v \in V$}{
            $\Omega_v \gets \texttt{mqtt.getDetectedObjects}(v)$\;
    }
    /* Periodically update the global object map */
    
    \If{$t-t_p \geq \tau $}{

         $\Omega_{all}  \gets \emptyset $;   /*  A unified detection list */ 
       
        \ForEach{CAV $v \in V$}{
            \ForEach{detected object $\omega \in \Omega_v$}{
                $\Omega_{all}$.\texttt{append}$(l_\omega, c_\omega, \omega.position)$\;
            }
        }
        
        /* Match detected objects to real locations */
        
        \ForEach{location $\rho \in \mathcal{R}$}{

            /* Find all detected objects within distance threshold $\delta$ of $\rho$ */

            $\mathcal{O}_\rho \gets \emptyset $;

            \ForEach{$\omega \in \Omega_{all}$}{
                \If{$\text{distance}(\omega.position, \rho) \leq \delta$}{
                    $\mathcal{O}_\rho$.\texttt{append}$(\omega)$\;
                 }
            }

            \If{$\mathcal{O}_\rho \neq \emptyset$}{ 
                /* Assign highest-confidence label */
                
                $l_\rho \gets \arg\max_{l} \sum_{\omega \in \mathcal{O}_\rho, l_\omega = l} c_\omega$;

                /* Compute the confidence score*/
                
                $c_\rho \gets \frac{\sum_{\omega \in \mathcal{O}_\rho} c_\omega^2}{\sum_{\omega \in \mathcal{O}_\rho} c_\omega}$;

                /* Compute the average position */
                
                $x_{\rho} \gets \frac{1}{|\mathcal{O}_\rho|} \sum_{\omega \in \mathcal{O}_\rho} x_{\omega}$;
                
                $y_{\rho} \gets \frac{1}{|\mathcal{O}_\rho|} \sum_{\omega \in \mathcal{O}_\rho} y_{\omega}$;
            }
            \Else{
                /* No object detected at $\rho$ */
                
                $l_\rho \gets \texttt{None}$\;
                $c_\rho \gets 0$\;
            }

        }
        /* Publish the updated global object map */

        $\Lambda \gets \{(\rho,l_\rho, c_\rho, x_\rho, y_\rho) \mid \forall \rho \in \mathcal{R}\}$\;
        \texttt{mqtt.publish}(``global\_detections'', $\Lambda$)\;

        $t_p \gets t$;
        
    }
}
\caption{PACE Edge Server Pseudocode}
\label{alg:PACE_Server}
\end{algorithm}

This approach is particularly useful in complex, multi-level, or highly obstructed environments where direct line-of-sight perception is often limited. By aggregating detections from multiple CAVs, PACE reduces individual sensor uncertainty and accelerates the the object labeling process, enabling faster and more accurate global perception updates.

\subsection{Variable Object Tally and Evaluation (VOTE)}
The goal of VOTE is to assess object labels in scenarios where many CAVs classify the same objects simultaneously, potentially assigning conflicting labels. VOTE facilitates a robust voting system to resolve discrepancies in diverse viewpoints and uncertain sensing environments by weighing the confidence scores of different labels based on CAV reputation and visibility parameters to generate a consensus label for each object. VOTE requires an unlabeled list of objects with known locations, which can be predefined (hard-coded) or detected with technologies such as LiDAR. The VOTE algorithm has two distinct components, the CAVs and the edge server.

\vspace*{0.2cm}
\noindent{\textbf{VOTE: CAV Component.}} 
Each CAV runs the VOTE client algorithm, which continuously scans its surroundings using a computer vision model~$\mathcal{M}$. The VOTE client algorithm follows the same structure as the PACE client, with minor modifications (the VOTE client pseudoscope is omitted for brevity). Each CAV subscribes to ``global\_verdicts'' via MQTT to receive finalized object labels, as VOTE focuses on label agreement. 
For each detected object~$\omega$, the CAV assigns a temporary label~$l_\omega$, a confidence score~$c_\omega$, and estimates its real position relative to its own frame of reference. The algorithm then cross-references detected objects with known object locations~$\mathcal{R}$, determining which real object each detection corresponds to based on spatial proximity. 
The CAV then packages the object's assigned temporary label, confidence, and estimated position, publishing the results to ``vote\_detections'' via MQTT to the edge server at regular intervals~$\tau$. 
The CAV component ensures that object detections from different viewpoints are consistently reported, enabling collaborative classification across multiple vehicles.

\vspace*{0.2cm}
\noindent{\textbf{VOTE: Edge Server Component.}}
The edge server component of VOTE is introduced in Algorithm~\ref{alg:VOTE_Server}. It can run on an edge server or any CAV acting as an aggregator, and it interacts with multiple CAVs simultaneously to process classification reports.
VOTE begins by initializing each CAV with a default reputation score, which corresponds to the proportion of correct labels issued by each CAV (lines~2-3).
Then, the algorithm creates an empty list of dictionaries, each corresponding to a known object location (lines~5-7). It then continuously processes incoming label reports from the CAVs (lines~8-21).

The collaborative decision-making process of VOTE is based on calculating an aggregated confidence score for each proposed label for each object using inputs from multiple CAVs (lines~11-13). VOTE considers three factors to calculate the aggregated confidence score: the reliability of the CAV, the confidence in the detected object label, and the visibility of the object from the CAV's perspective. 

The  aggregated confidence score for a given object at location~$\rho$ with a temporary detected label~$l$ (i.e., proposed label) is calculated as follows:
\begin{align}
%S_{\rho,l}&~=\sum_{v \in V}\sum_{\omega \in \Omega_v} f_l(\omega)~r_v~c_{\omega}~k(\rho,v)  \\
S_{\rho,l} &= \sum_{v \in V} \sum_{\omega \in \Omega_v} \underbrace{f_l(\omega)}_{\text{label match}} \cdot \underbrace{r_v}_{\text{reputation}} \cdot \underbrace{c_\omega}_{\text{confidence}} \cdot \underbrace{k(\rho, v)}_{\text{visibility}}
\label{eq:vote_final_score}
\end{align}
where~$V$~is the set of all CAVs, $\Omega_v$~is the set of all proposed object labels reported by CAV~$v$, $f_l(\omega)$~is an indicator function that equals~$1$ if $l_\omega = l$, and~0 otherwise,~$r_v$ is the reputation score of CAV~$v$, representing the historical reliability of its detections,~$c_\omega$ is the individual confidence score for the detected temporary label of object~$\omega$, and~$k(\rho,v)$ is the visibility score of the object at~$\rho$ relative to CAV~$v$. The visibility score~$k(\rho,v)$ accounts for both distance and angular positioning and is computed as:
\begin{align}
k(\rho,v)&~=~p_d~(1-\frac{d_{\rho v}}{d_{max}}) + (1-p_d)~(\frac{\theta_{\rho v}}{360\degree}) \label{eq:vote_visibility}
\end{align}
where~$p_d$ is a weight parameter that balances the impact of distance and angle on visibility,~$d_{\rho v}$ is the distance between location of the object~$\rho$ and the CAV~$v$,~$d_{max}$ is the maximum detection range, and~$\theta_{\rho v}$ is the absolute value of the angle created by a line segment from the CAV's camera to the object, relative to the CAV's camera orientation (representing deviation from the camera center). By incorporating both distance and angular clarity, VOTE prioritizes data from CAVs with better visibility of the object, ensuring that the final label decision is based on the most reliable observations.

At a specified verdict interval, VOTE determines the final consensus label~$\lambda_{\rho}$ for each object at~$\rho \in \mathcal{R}$ (line~18). 
\begin{align}
\lambda_{\rho}&~=~ \arg\max_{l} S_{\rho, l}
\label{eq:vote_verdict}
 \end{align}

\begin{algorithm}[t]
\KwData{List of actively-connected CAVs ($V$), global object location map ($\mathcal{R}$),  update interval ($\tau$)} 
\KwResult{Consensus object labels}

\texttt{mqtt.subscribe}(``vote\_detections''); /* Edge Server subscribes to CAV detections */

\ForEach{CAV $v \in V$}{
$r_v = 0.5$; /* Initial reputation score */ 
}

/* A nested dictionary to keep track of label votes */\\
$S$ = \text{new Dictionary};

\ForEach{location $\rho \in \mathcal{R}$}{
    $S$[$\rho$] $= \text{new Dictionary}$\;
}

\While{true}{
    /* Receive label votes from CAVs */\\
    \ForEach{CAV $v \in V$}{
        $\Omega_v \gets \texttt{mqtt.getDetectedObjects}(v)$\;
        \ForEach{$\rho,l$ in $\Omega_v$}{
            
            $S[\rho][l]~$+=$~( r_v~c_{\omega}~k(\rho,v))$\;
        }
    }
    
    /* Periodically determine a verdict (consensus label) */\\
    \If{$t-t_p \geq \tau$ }{
        
        /*Set the list of verdicts according to Eq.~\ref{eq:vote_verdict}*/\\

        $\Lambda \gets \emptyset$; /* Initialize verdicts set */

        \ForEach{location $\rho \in \mathcal{R}$}{
            /* Select label with highest confidence score */
            
            $\Lambda_\rho \gets \arg\max_{l} S[\rho][l]$;
            
        }

        /* Publish final consensus labels */
        
        \texttt{mqtt.publish}(``global\_verdicts'', $\Lambda$)\;

        /* Update CAV reputation scores */
        
    \ForEach{CAV $v$ in $V$}{
        $r_v \gets r_v + \Delta r_v$;    
    }

    /* Update last processed time */
        
     $t_p \gets t$;
    }
}
\caption{VOTE Edge Server Pseudocode}
\label{alg:VOTE_Server}
\end{algorithm}

After determining the final verdict, the edge server broadcasts the verdict to each CAV via an MQTT one-to-many query (line~19). This ensures that all CAVs update their local perception models with the agreed-upon object classifications.

To maintain fairness and improve the reliability of future decisions, the server adjusts each CAV's reputation score based on the correctness of its previous label contributions (lines~21-22). This reputation update parameter~$\Delta r_v$ is calculated as:
\begin{align}
\Delta r_v &= \texttt{cap}\left(\frac{\#~\text{correct} - \#~\text{incorrect}}{\#~\text{objects}},~30,~100\right)
\label{eq:vote_reputation}
\end{align}

The function ensures that reputation scores remain between 30 and 100, preventing extreme fluctuations and maintaining stability in trust levels. This adaptive reputation mechanism ensures that CAVs with a history of accurate detections gain more influence over future decisions, while unreliable CAVs contribute less to the consensus process. 

%The PACE and VOTE algorithms work to ensure reliable, real-time collaborative object detection, improving the situational awareness and decision-making of CAVs. 

Together, PACE and VOTE form the backbone of ECOD's collaborative perception framework. While PACE enhances situational awareness via multi-view fusion, VOTE ensures classification consistency through trust-aware consensus, jointly enabling real-time, accurate perception for CAVs.

\subsection{Computation and Communication Analysis}
In time-sensitive and bandwidth-constrained edge systems, such as autonomous vehicle networks, it is crucial for algorithms to maintain both low computational  and communication complexity.

The CAV components of both PACE and VOTE run in~$O(|\Omega_v|)$ time, where~$|\Omega_v|$ is the number of objects detected by the CAV. Most of the computation time on the client is attributed to the computer vision model, which can be independently optimized for autonomous vehicle applications. 
The edge server components of both algorithms run in $O(|V| \times |\Omega|)$ time, where $|V|$ is the number of CAVs connected to the server and $|\Omega|$ is the maximum number of objects labeled by each CAV (an upper bound on~$|\Omega_v|$). Note that for sufficiently small $|V|$,  the number of detected objects per CAV tends to be high, making~$|\Omega| \approx |\mathcal{R}|$, where $\mathcal{R}$ is the set of all known object locations. On the other hand, as $|V|$ increases, $|\Omega|$ may decrease due to visual obstructions between CAVs, leading to fewer detections per vehicle.

In terms of communication, ECOD minimizes bandwidth usage by transmitting only object-level summaries (labels, confidence scores, coordinates), unlike raw or feature-level fusion approaches. This makes the framework practical for bandwidth-constrained edge deployments.

Overall, VOTE and PACE scale very well in dynamic scenarios involving many vehicles and objects due to relatively low time complexity. Additionally, since all computationally expensive vision-related tasks are delegated to the CAVs, the edge server maintains a low processing load, ensuring real-time operation without introducing significant overhead.

\section{Experimental Results}
\label{experimentation}
This section describes the experimental setup and the experimental results of evaluating ECOD. 

\subsection{Experimental Setup}
We create a comprehensive experimental testbed consisting of four robotic CAVs equipped with cameras, an edge server, and a controlled test environment. 
The CAVs are built using the SunFounder Picar-X kit, each running on a Raspberry Pi~4 Model~B, the same hardware used for the edge server. The testbed is configured on a~60" by~40" grid divided into~2"x2" cells to standardize object placement and ensure consistent testing conditions across experiments.
Accuracy is defined as the proportion of correctly identified object detection verdicts over the course of multiple experimental cycles.

\vspace*{0.2cm}
\noindent{\textbf{Object Detection Tools.}}
Each CAV utilizes the object detection capability provided by the open-source Vilib computer vision library~\cite{vilib}, built on top of TensorFlow Lite~\cite{tensorflow2015-whitepaper} and OpenCV~\cite{opencv_library}. Vilib works in real time to detect and label objects from a live camera feed. 
Vilib is selected for compatibility with Raspberry Pi and testbed constraints. Our focus is on evaluating the collaborative gains, independent of specific object detector performance. 
Vilib's object detection module, trained on the Common Objects in Context (COCO) dataset~\cite{cocodataset}, can detect~80 classes of common household objects. The library also provides a QR code reader module, which is used in the PACE testbed. We extend Vilib's functionality by adding a feature that allows programmers to have more precise control over the object labels generated by the library's object detection and the QR reading module, integrating this feature into the onboard object detection module.

\vspace*{0.2cm}
\noindent{\textbf{Communication Setup.}}
We use MQTT to facilitate data transmission in real-time between CAVs and the edge server. The edge server runs the Eclipse Mosquitto MQTT broker (version~2.0.18)~\cite{mosquitto}. The CAVs and the edge server implement the Eclipse Paho MQTT library for Python integration~\cite{eclipse_paho_mqtt} to facilitate seamless real-time communication. 
The testbed relies on a stable local Wi-Fi network and MQTT for communication. While effective for prototyping, we note that real-world vehicular networks, such as LTE-V2X and 5G NR-V2X, exhibit variable latency, bandwidth constraints, and potential packet loss, which may impact performance. ECOD is compatible with these protocols and can be extended using SLAM/GNSS fusion. 
All components of the system run on Python~3.11.

\vspace*{0.2cm}
\noindent{\textbf{Benchmark.}}
We implement a single-CAV object detection method as a benchmark for comparison. In this baseline, each CAV independently detects nearby objects and reports its observations to the edge server without collaboration. This setup isolates the benefits of multi-vehicle coordination. Notably, the low  accuracy  of this benchmark is not due to shortcomings in the computer vision model, but rather due to occlusions, blind spots, and limited visibility inherent to single-vehicle sensing—challenges that ECOD is explicitly designed to overcome. 

\vspace*{0.2cm}
\noindent{\textbf{Localization.}}
For the purposes of our experiments, we assume that each vehicle knows its own global position, represented as a predefined two-dimensional Cartesian coordinate pair corresponding to a surface grid of two-inch squares.
In practice, this localization would ideally be determined by GPS or triangulation from nearby devices, as explored in prior work~\cite{liu2021edgesharing}. To simplify our experiments and control localization errors, we hard-code the global positions of the CAVs.

\begin{figure}[tbp]
    \centering
    {\includegraphics[width=0.85\linewidth]{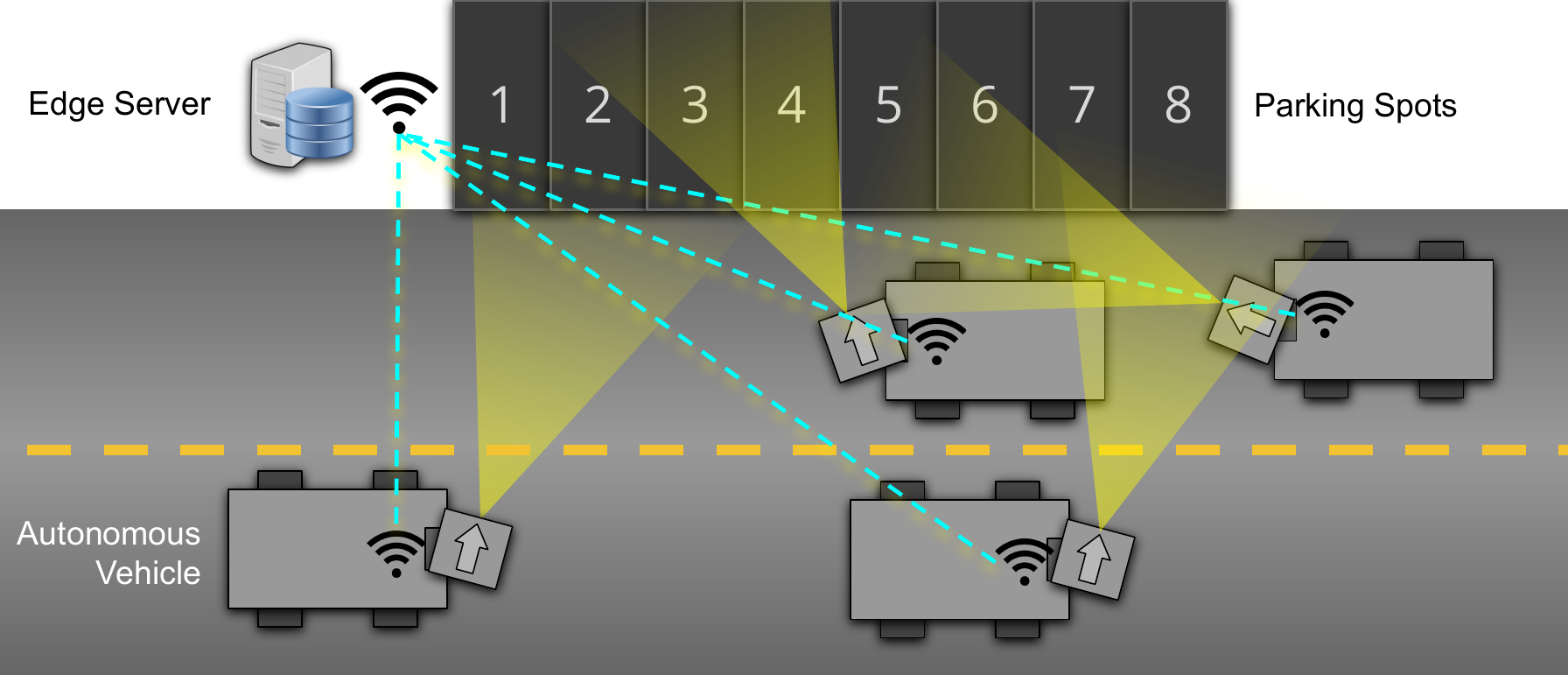}}
    \caption{PACE Testbed Diagram}
    \label{fig:pace_diagram}
\end{figure}

\begin{figure}[tbp]
    \centering
{\includegraphics[width=0.85\linewidth, height=0.25\textheight]{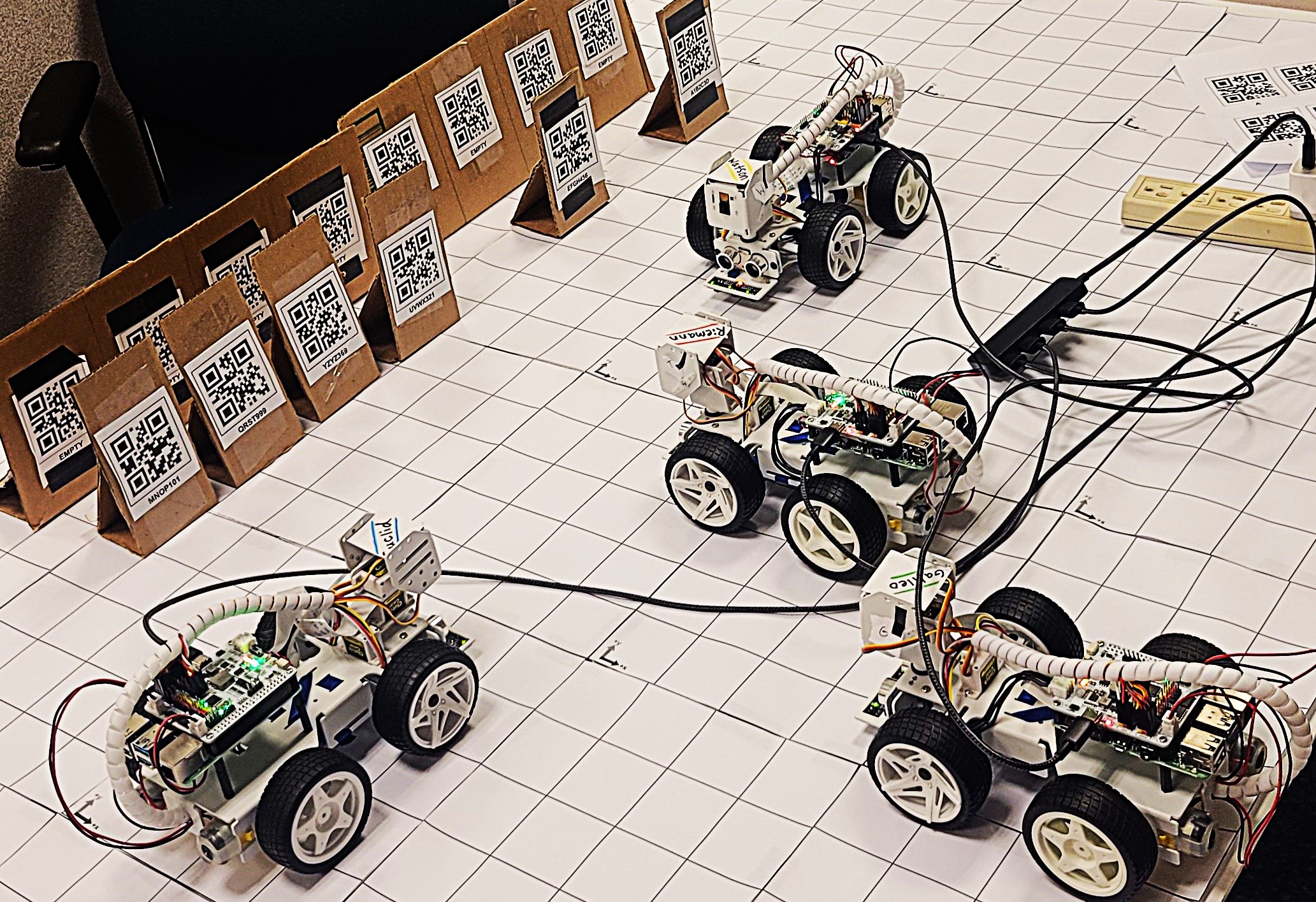}}
    \caption{PACE Testbed Photograph}
    \label{fig:pace_real_world}
\end{figure}

\vspace*{0.2cm}
\noindent{\textbf{PACE Testbed Setup.}}
To evaluate the PACE algorithm, we develop a parking lot testbed.  We utilize optical cameras (480p by 640p) and simulate parked vehicles using QR codes. 
Each QR code~(2.5" x~2.5") is mounted on a cardboard rectangle (2.75" wide by~5.5" tall) to represent a parked vehicle. These QR codes are arranged into one of six distinct parking configurations with varying levels of vehicle density. 
To ensure varied perspectives, the CAVs are positioned in three unique locations for each parking lot configuration. 
Figure~\ref{fig:pace_diagram} illustrates a diagram of the testbed, while Figure~\ref{fig:pace_real_world} displays a photograph of the setup for the four CAVs interacting with an edge server to collaboratively label the eight parking spots. 
For each distinct testbed configuration, we execute the experimental code, (available~\cite{main_github}), 
recording the outcomes of~200 sequential verdicts. We then calculate the average accuracy for PACE and the benchmark method.

\begin{figure}[tbp]
    \centering
    {\includegraphics[width=0.85\linewidth]{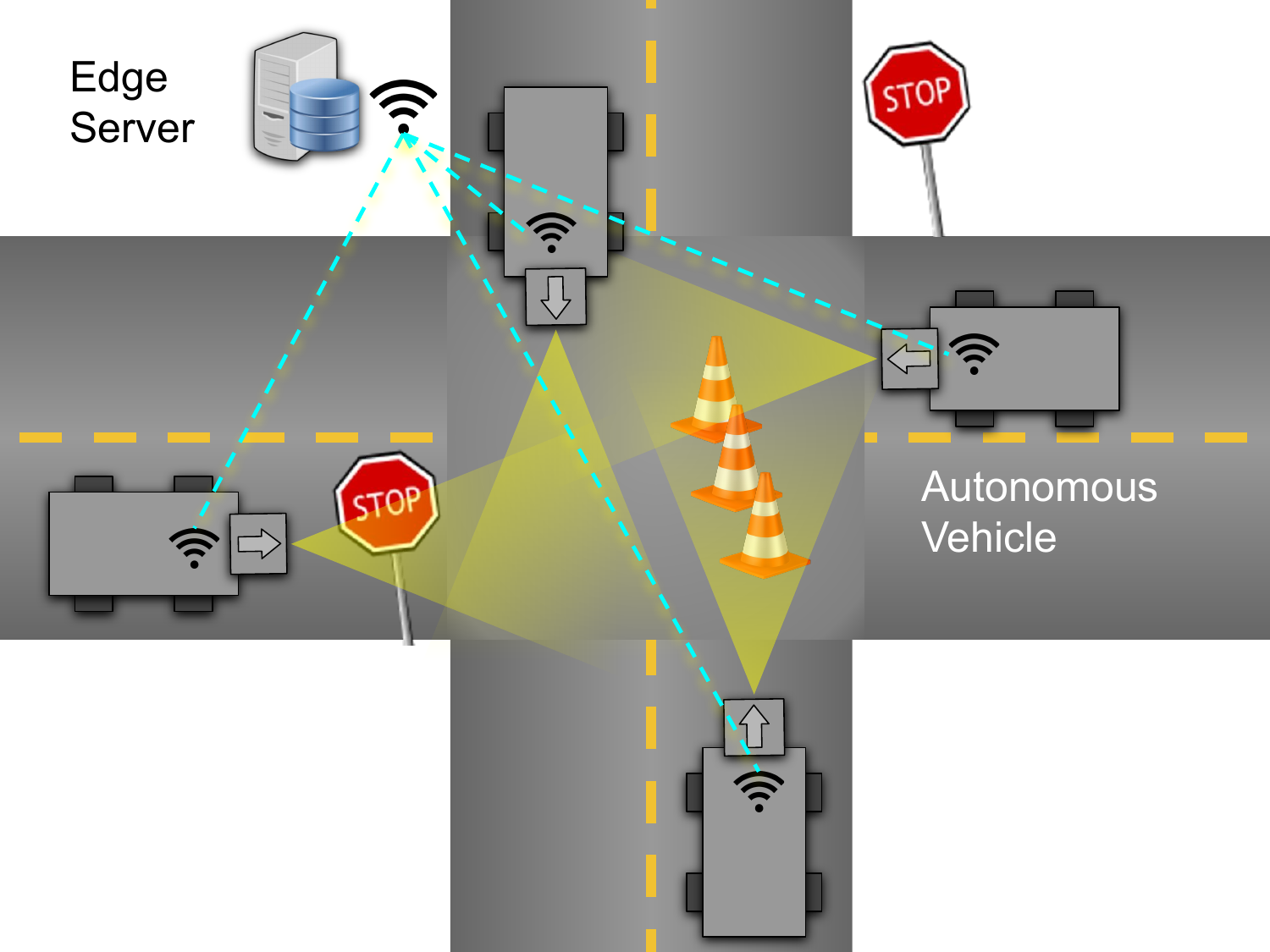}}
    \caption{VOTE Testbed Diagram}
    \label{fig:vote_diagram}
       % \vspace*{-0.2cm}
\end{figure}

\begin{figure}[tbp]
    \centering
    {\includegraphics[width=0.85\linewidth, height=0.25\textheight]{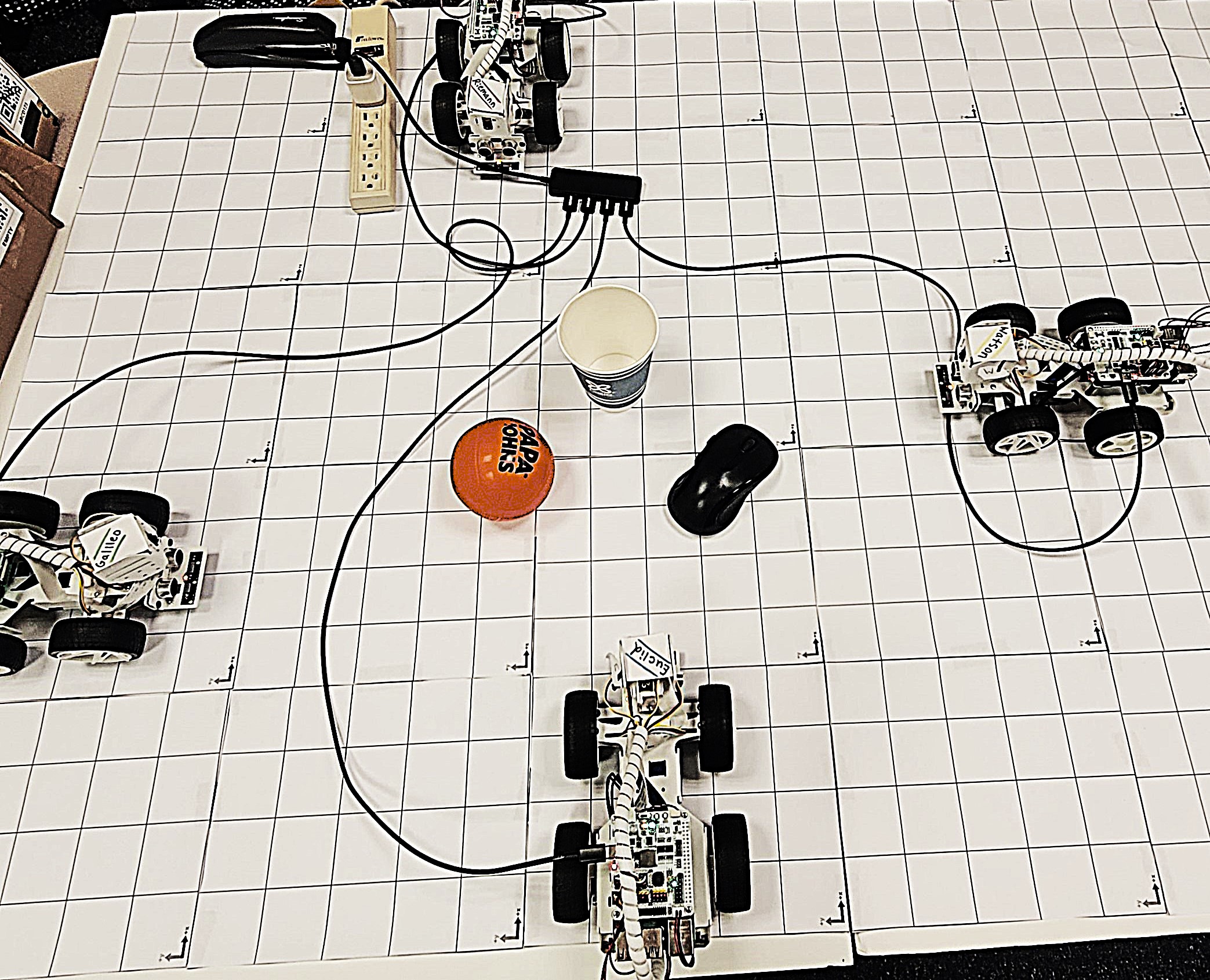}}
    \caption{VOTE Testbed Photograph}
    \label{fig:vote_real_world}
   % \vspace*{-0.3cm}
\end{figure}

\vspace*{0.2cm}
\noindent{\textbf{VOTE Testbed Setup.}}
To evaluate the VOTE algorithm, we develop a traffic intersection testbed. This setup contains three objects (a computer mouse, a solo cup, and an orange ball) placed at the center of the intersection. The CAVs view these objects from different angles, allowing for collaborative decision-making through voting. 
Each of these objects is discernible by our Vilib object detection model trained on the COCO dataset~\cite{vilib,cocodataset}. The relatively consistent detectability of these objects allows us to control the variable of onboard sensor efficacy, and instead focus on the efficacy of VOTE. We consider~0.5 default reputation score for each CAV to track their respective detection histories.
For flexibility when applying our server-side broker algorithm, we record all VOTE data as unprocessed object detection annotations from each CAV and then analyze the data using a post-synchronous adaptation of VOTE. For each of the three testbed configurations, we record~1,000 object detection cycles from each CAV, with verdicts being processed every~120ms. 
An example diagram of a VOTE testbed setup is illustrated in Figure~\ref{fig:vote_diagram}, while Figure~\ref{fig:vote_real_world} presents a photograph of an actual VOTE testbed setup.

\subsection{Analysis of Results} %\vspace*{-0.2cm}
\begin{figure}[tbp]
\vspace*{-0.2cm}
    \centering
    \includegraphics[width=0.80\linewidth]{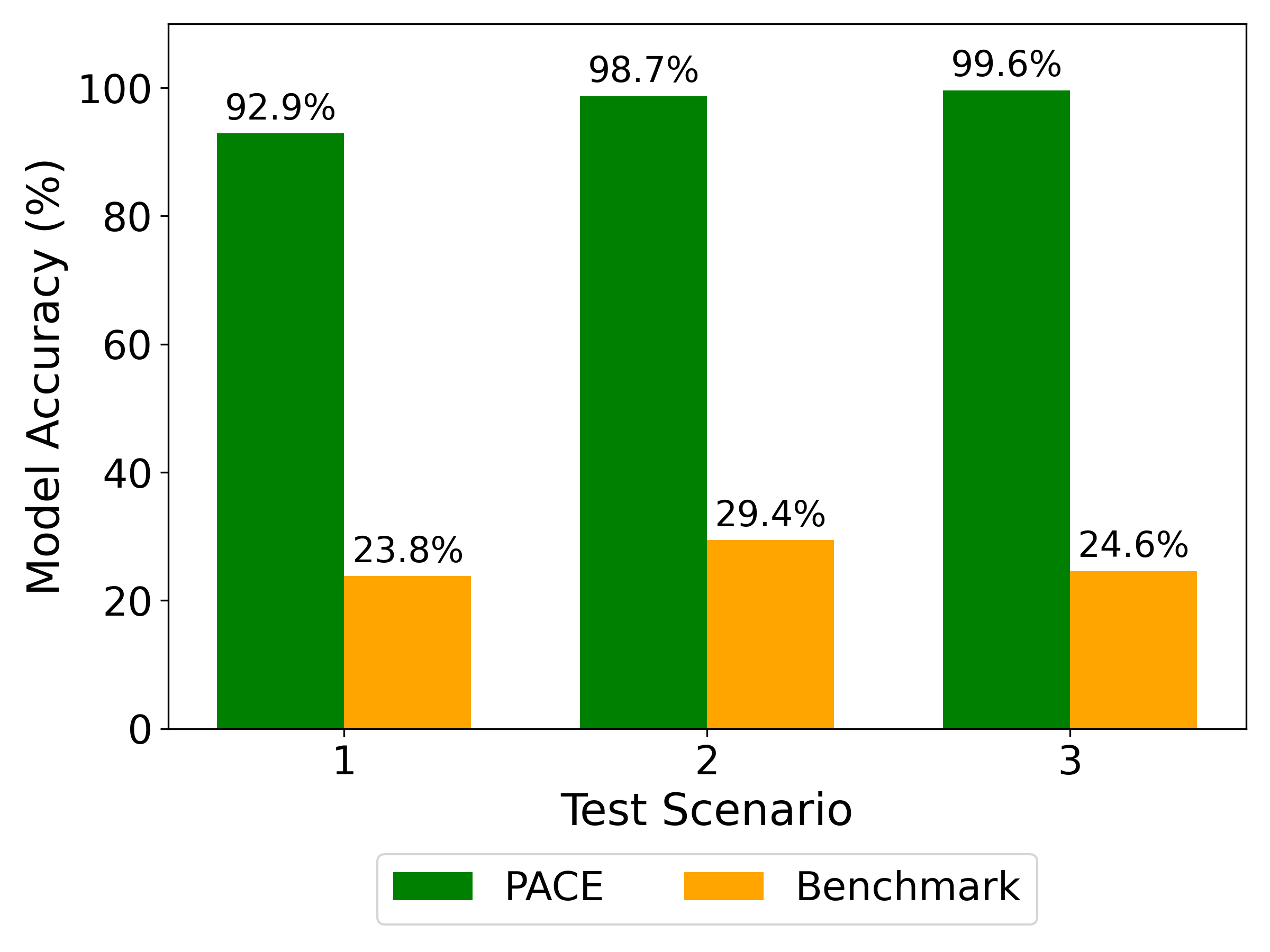}
    \caption{Model accuracy of PACE}%Comparison of model accuracy between PACE and the SCOD method, across three test CAVs configuration}
    \label{fig:pace_results}
    %\vspace*{-0.2cm}
\end{figure}

\begin{figure}[tbp]
    \centering
    \vspace*{-0.1cm}
    \includegraphics[width=0.80\linewidth]{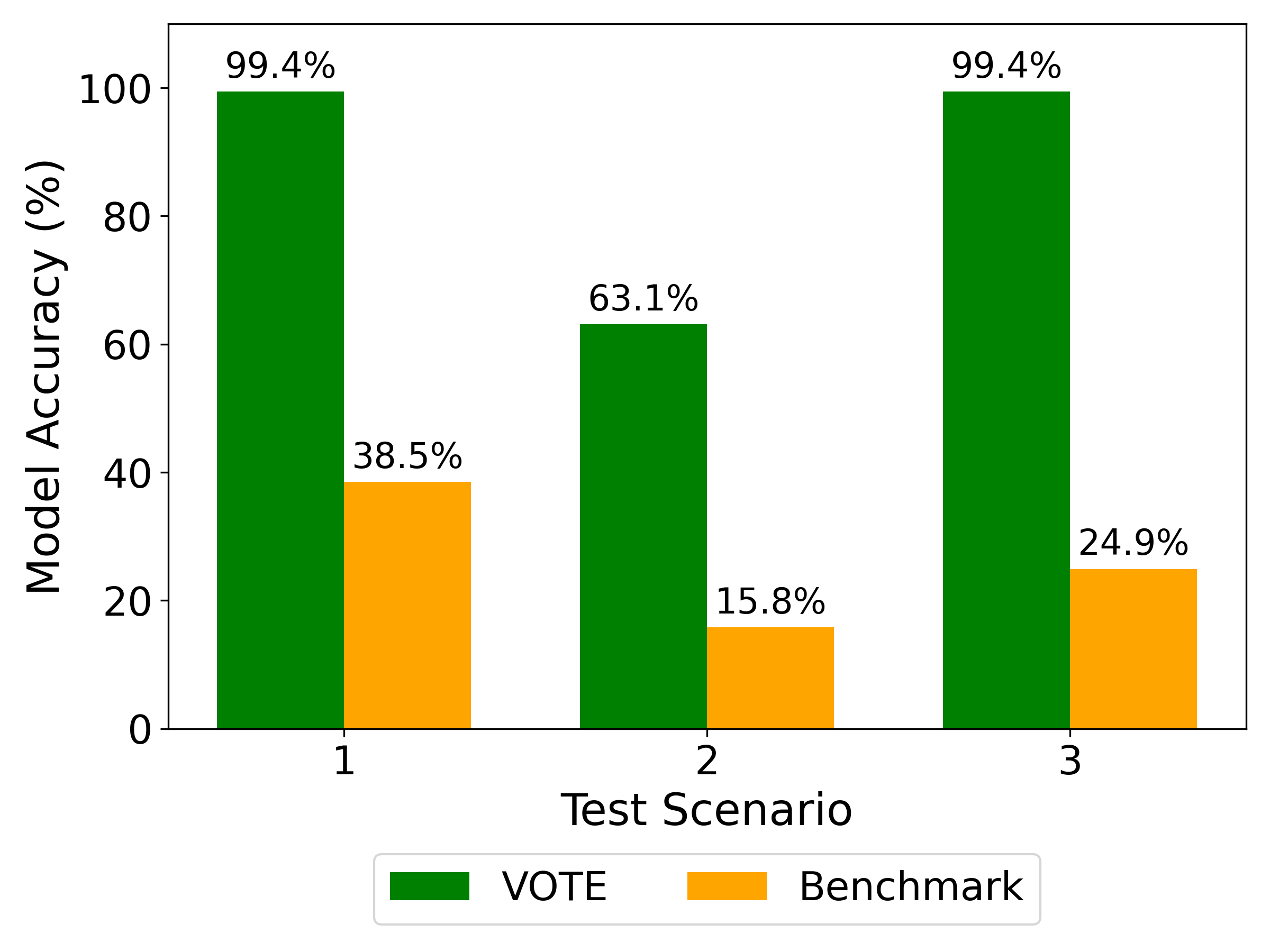}
    \caption{Model accuracy of VOTE}%Comparison of model accuracy between VOTE and the SCOD method, across three test objects}
    \label{fig:vote_results}
 % \vspace*{-0.2cm}
\end{figure}

Tables~\ref{table:pace_data} and~\ref{table:vote_data} summarize the results of PACE and VOTE, respectively. For each trial, the tables list the accuracy of the ECOD framework, the accuracy of the  benchmark, and the resulting improvement in accuracy (ECOD minus benchmark).  Figures~\ref{fig:pace_results} and~\ref{fig:vote_results} succinctly compare the accuracy of PACE and VOTE versus the  benchmark. 
Based on the results, PACE consistently outperforms the benchmark  in accuracy across all test configurations. On average, PACE achieves an accuracy of~97.1\%, compared to~25.9\% for the benchmark, reflecting a substantial improvement of~71.2\%. 
This improvement highlights PACE's effectiveness in enhancing object detection accuracy in scenarios where single-CAV perception is limited by occlusions and blind spots. 
This is attributed to the PACE algorithm's collaborative perception, which leverages multiple CAVs share their detections to achieve more accurate and reliable object detection.

Similarly,  VOTE  shows a marked improvement over the benchmark. VOTE achieves an average accuracy of~87.3\%, compared to~26.4\% for the benchmark, resulting in a~60.9\% improvement. 
This improvement reflects the VOTE algorithm's robust reputation-weighted  voting mechanism, which aggregates detections from multiple CAVs to resolve ambiguities and conflicts  in object labels. These results confirm the advantage of collaborative decision-making, especially in scenarios where multiple perspectives can contribute to a more accurate overall assessment.

In our prototype testbed, each perception-decision cycle completes in approximately every 100–150ms. This interval includes object detection on the CAV, MQTT transmission to the edge server, collaborative aggregation via PACE or VOTE, and publication of the consensus result. %While we did not explicitly measure round-trip latency, the system demonstrated sub-second responsiveness, suitable for real-time collaborative applications.}

\begin{table}
\caption{PACE Test Results} \vspace*{-0.1cm}
\label{table:pace_data}
\begin{center}
\def\arraystretch{1.5}
\scalebox{0.94}{
% \medium
\begin{tabular}{ |M{1.8cm}|M{1.8cm}|M{1.8cm}|M{1.8cm}|}
\hline
CAV Setup \# & PACE Accuracy (\%) & Benchmark Accuracy (\%) & Difference (\%) \\
\hline
1 & \textbf{92.9} & \textbf{23.8} & \textbf{+69.1}\\
\hline
2 & \textbf{98.7} & \textbf{29.4} & \textbf{+69.3}\\
\hline
3 & \textbf{99.6} & \textbf{24.6} & \textbf{+75.0}\\
\hline
\hline
All trials & \textbf{97.1} & \textbf{25.9} & \textbf{+71.2}\\
\hline
\end{tabular}
}
\vspace{0.1cm}
\end{center}
%\vspace*{-0.2cm}
\end{table}

% \begin{table}
% \def\arraystretch{1.5}
% \scalebox{1.0}{
% % \medium
% \begin{tabular}{ |M{0.5cm}|M{0.9cm}|M{1.8cm}|M{1.8cm}|M{1.0cm}|}
% \hline
% Trial \# & Object Config & VOTE Accuracy (\%) & Benchmark Accuracy (\%) & Diff. (\%) \\

% \hline
% 1 & I & 99.000 & 38.042 & +60.958\\
% \hline
% 2 & I & 99.500 & 38.583 & +60.917\\
% \hline
% 3 & I & 99.667 & 38.875 & +60.792\\
% \hline
% \multicolumn{2}{|c|}{Avg. trials 1-3} & \textbf{99.389} & \textbf{38.500} & \textbf{+60.889}\\
% \hline
% 4 & II & 61.333 & 15.333 & +46.000\\
% \hline
% 5 & II & 63.500 & 15.875 & +47.625\\
% \hline
% 6 & II & 64.333 & 16.083 & +48.250\\
% \hline
% \multicolumn{2}{|c|}{Avg. trials 4-6} & \textbf{63.055} & \textbf{15.764} & \textbf{+47.292}\\
% \hline
% 7 & III & 99.167 & 24.792 & +74.375\\
% \hline
% 8 & III & 99.667 & 24.917 & +74.750\\
% \hline
% 9 & III & 99.333 & 24.833 & +74.500\\
% \hline
% \multicolumn{2}{|c|}{Avg. trials 7-9} & \textbf{99.389} & \textbf{24.847} & \textbf{+74.542}\\
% \hline
% \hline
% \multicolumn{2}{|c|}{Avg. all trials} & \textbf{87.278} & \textbf{26.370} & \textbf{+60.907}\\
% \hline
% \end{tabular}
% }
% \vspace{0.1cm}
% \caption{VOTE Test Results}
% \label{table:vote_data}
% \end{table}

\begin{table}
\caption{VOTE Test Results} \vspace*{-0.1cm}
\label{table:vote_data}
\begin{center}
\def\arraystretch{1.5}
\scalebox{0.94}{
% \medium
\begin{tabular}{ |M{1.8cm}|M{1.8cm}|M{1.8cm}|M{1.8cm}|}
\hline
CAV Setup \# & VOTE Accuracy (\%) & Benchmark Accuracy (\%) & Difference (\%) \\
\hline
1 & \textbf{99.4} & \textbf{38.5} & \textbf{+60.9}\\
\hline
2 & \textbf{63.1} & \textbf{15.8} & \textbf{+47.3}\\
\hline
3 & \textbf{99.4} & \textbf{24.8} & \textbf{+74.5}\\
\hline
\hline
All trials & \textbf{87.3} & \textbf{26.4} & \textbf{+60.9}\\
\hline
\end{tabular}
}
%\vspace{0.1cm}
\end{center}
\vspace*{-0.1cm}
\end{table}

The experimental results achieved by both PACE and VOTE validate the effectiveness of  ECOD  in significantly improving object detection accuracy and decision-making in complex, dynamic environments. By leveraging collaborative perception and real-time data aggregation through edge computing, the ECOD framework overcomes the key limitations of single-CAV perception, demonstrating its potential to enhance  safety and reliability of autonomous vehicle systems.

\section{Conclusion}
\label{conclusion} 
We introduced the ECOD framework, an edge-enabled collaborative object detection framework for CAVs, featuring two novel algorithms, PACE and VOTE. 
PACE enables efficient, real-time object classification by leveraging cooperative perception in occluded or complex environments, while VOTE facilitates consensus-based label agreement among multiple CAVs via confidence-weighted voting, enhancing detection reliability. 
Experimental evaluations on a multi-CAV testbed  demonstrated that both algorithms significantly enhance object detection accuracy compared to a traditional single-CAV benchmark, validating the benefits of collaborative perception in autonomous systems. By advancing edge-enabled cooperation, ECOD contributes to the broader goal of safer and more reliable autonomous vehicle systems, paving the way for scalable, intelligent perception frameworks in next-generation smart mobility. 
These findings show the potential of  distributed cooperative systems, where perception and decision-making are shared across many intelligent vehicles at the edge, reducing reliance on centralized infrastructure and enhancing robustness. 
Future work will focus on improving system scalability, optimizing computational efficiency for real-time inference, and incorporating adaptive learning mechanisms to refine object classification over time.  We aim to expand ECOD's applicability to more challenging environments, such as high-traffic intersections and urban driving scenarios, with mixed vehicles where dynamic collaboration can significantly enhance both safety and efficiency. We also seek to explore adaptive collaboration strategies, where CAVs dynamically adjust their participation based on visibility, confidence, and task priorities.

%\textcolor{red}{\st{Future work will focus on scaling ECOD to real-world CAV deployments by improving system scalability, optimizing real-time inference, and integrating multimodal sensors such as LiDAR and radar. We also plan to incorporate adaptive learning mechanisms for dynamic trust modeling and object classification. To broaden ECOD’s applicability, we aim to test in more complex environments—such as high-traffic intersections and urban roads with mixed vehicle types—where collaborative perception can significantly improve safety and efficiency. Finally, we will explore adaptive coordination strategies, enabling CAVs to adjust their participation based on visibility, confidence, and task relevance.}}

\vspace*{0.2cm}
\noindent{\textbf{Acknowledgments.}}
This work was supported in part by the National Science Foundation under Grants No. 2050879 and 2145268.

\bibliographystyle{IEEEtran}
\bibliography{references}

\end{document}